\documentclass{article}

\usepackage[preprint, nonatbib]{neurips_2023}

\usepackage{longtable}
\usepackage[utf8]{inputenc} 
\usepackage[T1]{fontenc}   
\usepackage{hyperref}      
\usepackage{url}            
\usepackage{booktabs}       
\usepackage{amsfonts}       
\usepackage{CJKutf8}
\usepackage{nicefrac}       
\usepackage{microtype}      
\usepackage{xcolor}        
\usepackage{amsmath}
\usepackage[most]{tcolorbox}
\usepackage{makecell} 
\usepackage{array}    
\usepackage{float}
\usepackage{graphicx}
\usepackage{ltablex}
\usepackage{lipsum}
\usepackage{graphicx}
\usepackage{wrapfig}
\usepackage[size=small]{caption}
\usepackage{mathtools}
\usepackage{tabularx}
\usepackage{amssymb}
\usepackage{amsthm}
\usepackage{soul}
\usepackage{float}
\usepackage{xspace}
\definecolor{textblue}{rgb}{.2,.2,.7}
\definecolor{textred}{rgb}{0.54,0,0}
\definecolor{textblack}{rgb}{0,0,0}
\definecolor{textgreen}{rgb}{0,0.53,0}
\usepackage{listings}
\lstdefinestyle{pythonstyle}{
    language=Python,
    morekeywords={None}, % Add any keywords you want to highlight
    breaklines=true,
}

\definecolor{codegreen}{rgb}{0,0.6,0}
\definecolor{codegray}{rgb}{0.5,0.5,0.5}
\definecolor{codepurple}{rgb}{0.58,0,0.82}
\definecolor{backcolour}{rgb}{0.95,0.95,0.92}

\lstdefinestyle{mystyle}{
    backgroundcolor=\color{backcolour},   
    commentstyle=\color{codegreen},
    keywordstyle=\color{magenta},
    numberstyle=\tiny\color{codegray},
    stringstyle=\color{codepurple},
    basicstyle=\ttfamily\footnotesize,
    breakatwhitespace=false,         
    breaklines=true,                 
    captionpos=b,                    
    keepspaces=true,                 
    numbers=left,                    
    numbersep=5pt,                  
    showspaces=false,                
    showstringspaces=false,
    showtabs=false,                  
    tabsize=2
}

\usepackage{array}
\newcolumntype{M}[1]{>{\centering\arraybackslash}m{#1}}

\title{Octo-planner: On-device Language Model for Planner-Action Agents}

\author{%
  Wei Chen\thanks{Equal contribution.}\\
  Nexa AI \& Stanford\\
  Sunnyvale, CA 94086 \\
  \texttt{alexchen@nexa4ai.com} \\
  \And
  Zhiyuan Li$^*$\\
  Nexa AI \& Stanford\\
  Sunnyvale, CA 94086 \\
  \texttt{zack@nexa4ai.com}
  \AND
  \hspace{0.5cm}Zhen Guo$^*$ \\
  \hspace{0.5cm}MIT EECS\\
  \hspace{0.5cm}Cambridge, MA 02139 \\
  \hspace{0.5cm}\texttt{zguo0525@mit.edu} \\
  \And
  Yikang Shen$^*$ \\
  MIT-IBM Watson AI Lab \\
  Cambridge, MA 02139 \\
  \texttt{yikang.shen@ibm.com} \\
}

\begin{document}
% \begin{CJK*}{UTF8}{gbsn}
% \nocite{*}
\maketitle

\begin{abstract}
AI agents have become increasingly significant in various domains, enabling autonomous decision-making and problem-solving. To function effectively, these agents require a planning process that determines the best course of action and then executes the planned actions. 
In this paper, we present an efficient on-device Planner-Action framework that separates planning and action execution into two components: a planner agent, or Octo-planner, optimized for edge devices, and an action agent using the Octopus model for function execution. Octo-planner first responds to user queries by decomposing tasks into a sequence of sub-steps, which are then executed by the Octopus action agent. To optimize performance on resource-constrained devices, we employ model fine-tuning instead of in-context learning, reducing computational costs and energy consumption while improving response times. Our approach involves using GPT-4 to generate diverse planning queries and responses based on available functions, with subsequent validation to ensure data quality. We fine-tune the Phi-3 Mini model on this curated dataset, achieving a 97\% success rate in our in-domain test environment. To address multi-domain planning challenges, we develop a multi-LoRA training method that merges weights from LoRAs trained on distinct function subsets. This approach enables flexible handling of complex, multi-domain queries while maintaining computational efficiency on resource-constrained devices. To support further research, we have open-sourced our model weights at \url{https://huggingface.co/NexaAIDev/octopus-planning}. For the demo, please refer to \url{https://www.nexa4ai.com/octo-planner#video}.
\end{abstract}

\begin{figure}[ht!]
  \centering
  \includegraphics[width=0.7\textwidth]{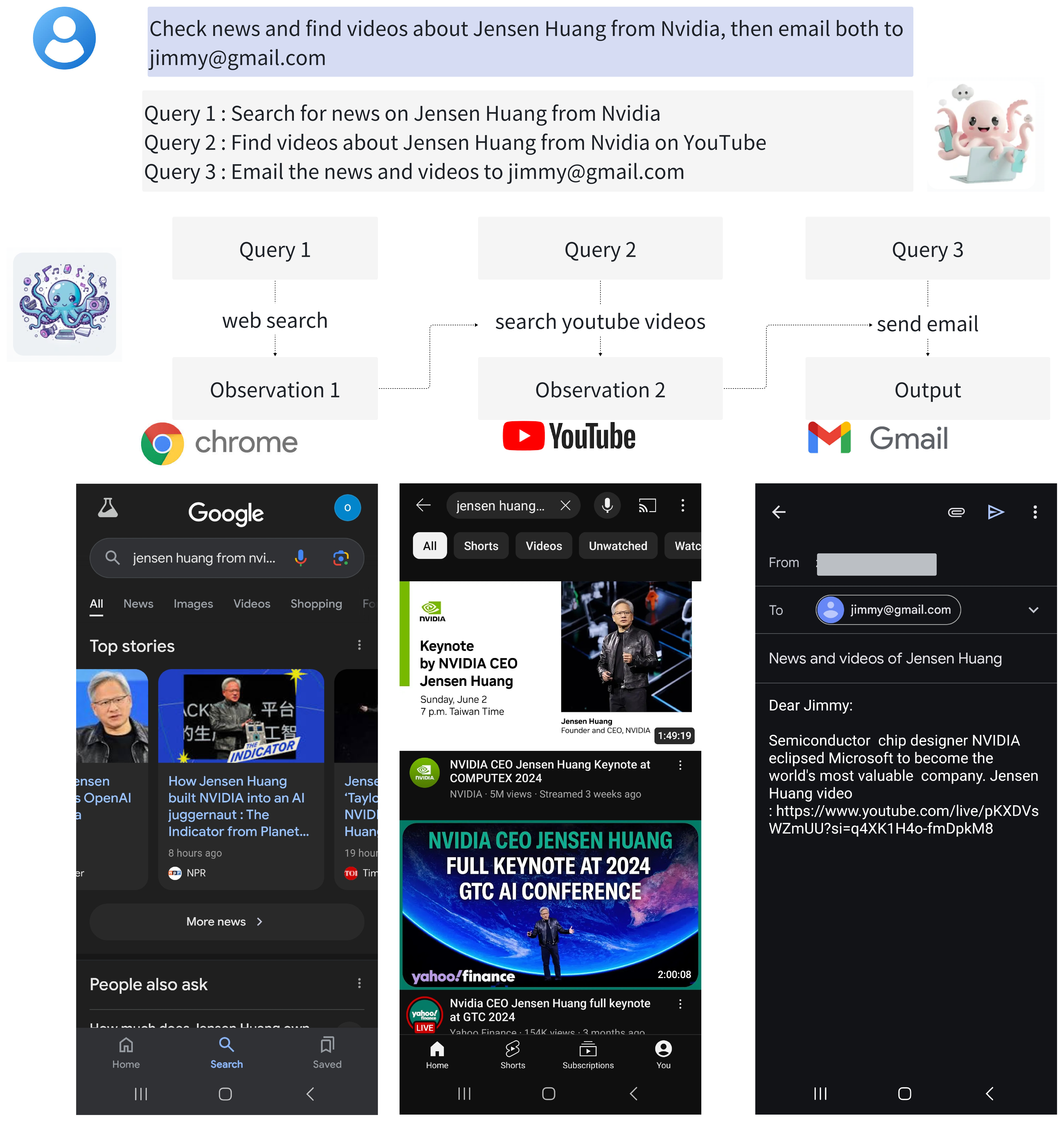}
  \caption{Planner-Action Agent on a smartphone using Octopus models}
\end{figure}

\section{Introduction}
Artificial intelligence (AI) agents~\cite{jennings1998applications, poole2010artificial} have significantly transformed various industries by enabling autonomous decision-making and improving operational efficiencies~\cite{kim2023language, deng2023mind2web, yan2023gpt4v, zheng2024gpt4vision, koh2024visualwebarena}. These agents rely on a critical planning process that involves determining the optimal course of action, executing the planned actions, and summarizing the outcomes. Large Language Models (LLMs) such as Gemini-Pro~\cite{geminiteam2024gemini} and GPT-4~\cite{openai2024gpt4} have shown potential in this domain. While these models face challenges in executing complex planning tasks at a level comparable to human performance~\cite{xie2024travelplanner, zheng2024natural}, they remain effective in addressing simpler tasks, thereby facilitating practical applications. 
One such application is the emergence of AI assistant tools from companies like MultiOn~\cite{multion2024}, Simular AI~\cite{simular2024}, and Adept AI~\cite{adept2024}, which leverage the capabilities of LLMs to provide intelligent assistance across various domains. Additionally, consumer-oriented AI hardware products, such as Rabbit R1~\cite{rabbit2024}, Humane AI Pin~\cite{humane2024}, and Limitless Pendant~\cite{limitless2024}, integrate LLMs into user-friendly devices, making intelligent assistance more accessible and gaining significant traction.

The success of AI agents depends on the performance of the underlying LLMs. Agents using pre-trained models without fine-tuning on task demonstrations have relatively low success rates, ranging from 12\% on desktop applications~\cite{xie2024osworld} to 46\% on mobile applications~\cite{bishop2024latent}, while those leveraging fine-tuned models can achieve success rates of up to 80\% on tasks similar to their training data~\cite{nakano2022webgpt, gur2024realworld}. However, using LLMs for AI agents is costly due to high computational demands and infrastructure expenses, which limits widespread adoption. The lack of on-device AI agents restricts applications requiring real-time processing, offline functionality, or enhanced privacy.

On-device AI agents offer advantages including reduced latency, offline operation, lower costs, and improved data security~\cite{yu2017survey, lin2024awq, alwarafy2020survey, ranaweera2021survey}. While action models like Octopus V2 achieve over 95\% accuracy in function calling~\cite{chen2024octopus2}, an on-device planning model is still missing. General agent frameworks rely on single-model in-context learning, which requires lengthy function descriptions and planning instructions in each prompt. This approach is impractical for on-device models with limited context lengths, as it causes high latency and battery consumption on edge devices.

In this paper, we introduce Octo-planner, an on-device planning agent that addresses the key challenges of efficiency, adaptability, and resource constraints. Our Planner-Action framework separates planning and action execution into two components: a planner agent, or Octo-planner, optimized for edge devices, and an action agent using the Octopus model for function execution. By prioritizing fine-tuning over few-shot prompting, we reduce computational costs and minimize key-value (KV) cache requirements. Our approach uses GPT-4 to generate and validate planning data, which is then used to fine-tune Phi-3 Mini for on-device deployment. In-domain tests demonstrate that this fine-tuning improves planning success rates to 97\%. To address multi-domain planning challenges, we develop a multi-LoRA training method that merges weights from LoRAs trained on distinct function subsets. This method enables flexible handling of complex, multi-domain queries while maintaining computational efficiency on resource-constrained devices. By focusing on pre-defined functions for simpler tasks and leveraging fine-tuning, we aim to make AI agents more practical, accessible, and cost-effective for real-world applications.

This work aims to contribute to the ongoing efforts to make AI more accessible and practical for everyday use. By bridging the gap between AI agent potential and edge computing constraints, we seek to facilitate the adoption of intelligent, on-device assistants across various domains. Through open-sourcing our approach, we hope to inspire further innovations in on-device AI, expanding the reach of advanced planning capabilities to a broader range of applications.

\section{Related Work}
% In this section, we outline the related works about our current project. We intend to use fine-tune to replace the prefix prompting to achieve high accuracy, low latency and low energy consumption. 

\textbf{Planner agent}\quad Language models have become essential in planning agent systems. Proprietary models like OpenAI's assistant API~\cite{openai_assistant_overview} excel in generating strategies based on user queries and available functions. %These models show enhanced performance in interactive and embodied environments, particularly when fine-tuned on domain-specific data and tasks.
Recent advancements have further expanded the capabilities of language models in planning. The ReAct framework~\cite{yao2023react} integrates planning and acting for limited action spaces, while research from Alibaba Group~\cite{shen2024small} highlights the effectiveness of separate planning and action models for complex tasks. In robotics, language models are also increasingly applied to task-level planning~\cite{hu2023generalpurpose, firoozi2023foundation}. Notable examples include SayCan~\cite{ahn2022i}, which uses LLMs to break high-level tasks into concrete sub-tasks, and Video Language Planning (VLP)~\cite{du2023video}, which enhances long-horizon planning through a text-to-video dynamics model.
The broad application of language models in planning systems, from general strategies to specific robotics tasks, underscores their growing importance and adaptability in decision-making processes across diverse domains.

\textbf{Fine-tuning to replace long context}\quad Fine-tuning language models to internalize specific prompts or context information reduces input length and improves efficiency~\cite{lester2021power, li2021prefix}. This approach involves training models on carefully curated, task-specific datasets. For models with limited context windows, this technique is particularly valuable as it enables more efficient query processing without sacrificing response quality. The success of fine-tuning largely depends on the use of diverse, high-quality datasets, which ensure the model can generalize across various prompt phrasings~\cite{paul2023deep, cao2023instruction, xia2024less, wang2024diversity}. When implemented effectively, fine-tuning streamlines application-specific interactions, addressing both context length limitations and computational challenges in practical deployments.

\textbf{LoRA and Multi-LoRA}\quad Low-Rank Adaptation (LoRA) efficiently adapts pre-trained language models to specific tasks~\cite{hu2021lora}. Unlike full fine-tuning, which updates all parameters, LoRA freezes the pre-trained weights and adds trainable low-rank matrices to each layer, significantly reducing the number of trainable parameters and the computational demands. Multi-LoRA extends this concept by enabling multiple task-specific adapters to be trained, combined, or switched during inference, allowing a single base model to handle various tasks efficiently~\cite{wang2023multilora}. Building on these approaches, researchers have developed several related variants to address different aspects of model adaptation: LoRA+ optimizes learning rates~\cite{hayou2024lora}, VeRA uses random projections~\cite{kopiczko2024vera}, AdaLoRA implements adaptive rank~\cite{zhang2023adalora}, DoRA decomposes weights~\cite{liu2024dora}, and Delta-LoRA updates the pre-trained weights~\cite{zi2023deltalora}. These variants aim to further refine efficiency or performance in specific scenarios.

%LoRA and Multi-LoRA are particularly useful for large models and resource-constrained scenarios, offering flexibility across multiple tasks. The choice between these methods depends on data availability, computational resources, and task requirements.

\section{Method}
This section presents our framework for on-device Planner-Action agents. We first describe the integration of planning and action agents for efficient problem-solving. We then detail our approach to dataset design and the training process for the planning agent, including support for extensive function sets and a plug-and-play capability for additional ones. Finally, we outline the benchmark used to evaluate agent performance.

\subsection{Planner and action agents framework}
Our Planner-Action approach distinguishes itself from general agent frameworks by separating the planning and action execution processes into two components. This separation improves modularity and allows for specialized optimization of each component. The framework operates as follows: 

\textbf{Planner Phase}: Given a user query $q$, our planning model $\pi_{plan}$ decomposes the task into a sequence of sub-steps. Formally:
\begin{equation}
\{\tau_1, \tau_2, ..., \tau_n\} = \pi_{plan}(q; F),
\end{equation}
where $F$ is the set of available function descriptions, and $\tau_i$ is the $i^{th}$ execution step. $\pi_{plan}$ internalizes $F$ during instruction fine-tuning.

\textbf{Action Phase}: For each step in the execution sequence, we employ an action model $\pi_{action}$. At step $i$, given the observation of the current state $O_i$, the action model performs:
\begin{equation}
O_{i+1} = \pi_{action}(\tau_i, O_i),
\end{equation}
where $O_{i+1}$ and $\tau_{i+1}$ are passed to the next step for continued execution. This iterative process ensures coherent progression through the task's sub-steps. 

\begin{figure}[ht]
    \centering
    \includegraphics[width=\textwidth]{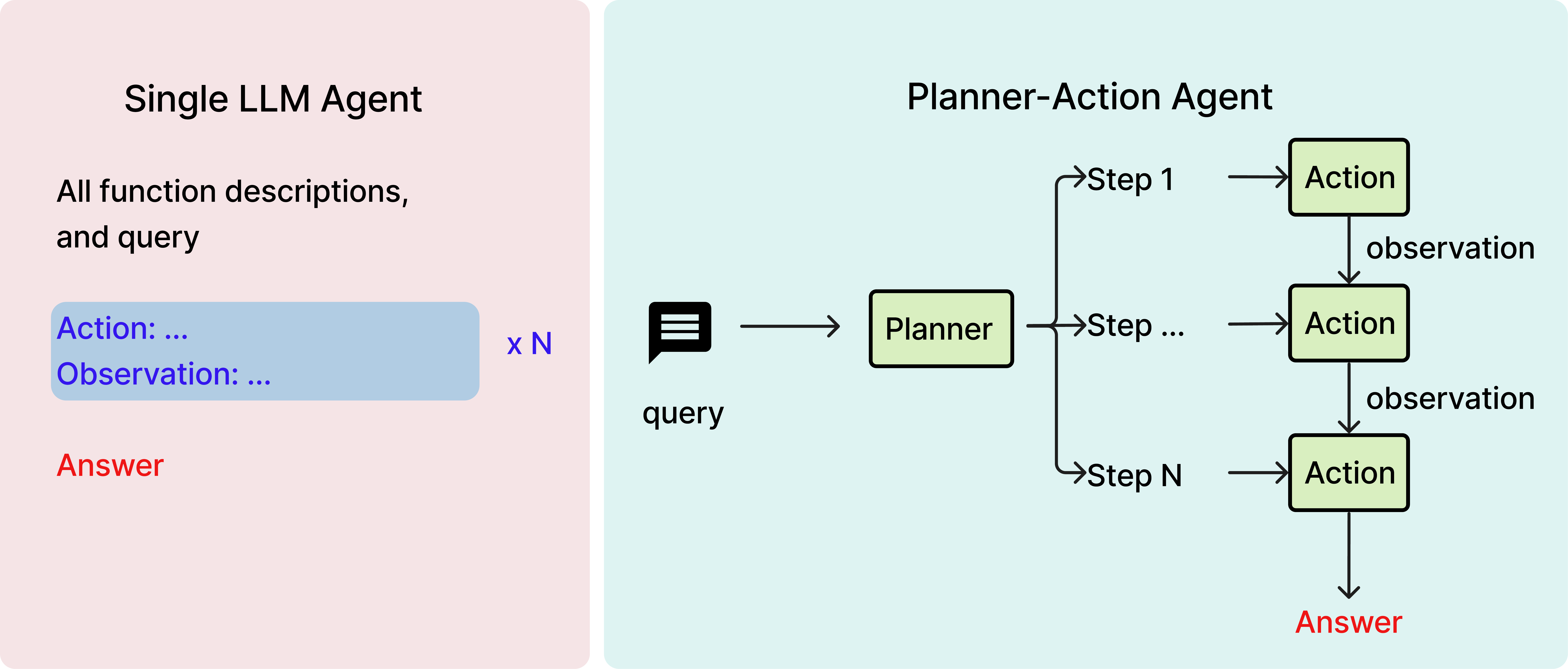}
    \caption{Comparison of Single LLM Agent and Planner-Action Agent frameworks. 
    (left) Single LLM Agent: A unified model performs both task planning and action execution. 
    (right) Planner-Action Agent: A specialized planner model decomposes the task into subtasks, 
    while a separate action model executes each subtask sequentially.}
    \label{fig:multiagent}
\end{figure}

For the action model, we utilize the Octopus model, which is specifically designed for on-device function calling. Figure \ref{fig:multiagent} illustrates the difference between our Planner-Action framework and the single-model approach for LLM agents. 
%Our framework's modular design allows for more efficient task decomposition and specialized execution, potentially leading to improved performance and flexibility in complex scenarios.
%Our framework makes planning and action using purely small on-device model possible. This comparison also highlights the potential advantages of our modular architecture in terms of flexibility, scalability, and targeted optimization.

The modular design of our framework offers several advantages:
\begin{itemize}
\item \textbf{Specialization}: Separating planning and action execution allows optimization of each model for its specific role, enhancing performance in complex tasks.
\item \textbf{Scalability}: Independent scaling of planning and action capabilities efficiently accommodates varying task complexities.
\item \textbf{Interpretability}: Explicit separation of phases improves transparency in the decision-making process.
\item \textbf{Adaptability}: Domain-specific knowledge or constraints can be integrated into either phase without system-wide changes.
\end{itemize}

% Subsequent sections detail our dataset design, training methods, and evaluation metrics.

\subsection{Planning dataset}
Our framework uses the Octopus model as the action model, requiring training only for the planner agent. We fine-tune the planner agent with the following dataset format:
\begin{tcolorbox}
<|user|>\{user's query\}<|end|>\\
\\
<|assistant|> \texttt{\{query$_1$\}}<nexa\_split>\texttt{\{query$_2$\}}<nexa\_split>...<nexa\_split>\texttt{\{query$_n$\}}.<|end|>
\end{tcolorbox}

Special tokens such as \texttt{<|user|>} and \texttt{<|assistant|>} are used for chat model pre-training but are optional otherwise. We set $n$ to range from 1 to 5, based on our finding that most tasks on mobile apps consist of fewer than 5 steps. The dataset generation and curation process includes:

\begin{enumerate}
    \item \textbf{Dataset collection}: Given the available functions $F$, we use a large language model (GPT-4) to generate diverse queries answerable by these functions. We increase the model's temperature setting to ensure query variety. Responses are then generated in the specified dataset format. Importantly, while function descriptions are used during generation, they are not included in the final dataset. Instead, the planner model internalizes this function information during training.
    \item \textbf{Data validation}: We employ the same language model as a validation tool to assess the correctness of query-response pairs. Despite some errors in the initial generation process, we found that the model effectively classifies generated content as valid or invalid, allowing us to filter out incorrect outputs and maintain dataset quality.
\end{enumerate}

Example data points with different numbers of sub-steps are shown below:

\begin{tcolorbox}
\textbf{\fontsize{11}{13}\selectfont One-step example}:\\

\textbf{query}: Tell me today's stock market.

\textbf{response}: Search for today's stock market news and latest updates<|end|>\\

\textbf{\fontsize{11}{13}\selectfont Two-step example}:\\

\textbf{query}: Find and email me Jensen Huang's news in English.

\textbf{response}: Find articles about Jensen Huang in English<nexa\_split> Email the first article found to myself<|end|>\\

\textbf{\fontsize{11}{13}\selectfont Three-step example}:\\

\textbf{query}: Find and increase the screen brightness by 20\% to better view the quarterly report document before connecting to the office printer via Bluetooth.

\textbf{response}: Find the quarterly report document in the system<nexa\_split> Increase screen brightness by 20\% to improve visibility<nexa\_split> Connect to the office printer via Bluetooth to print the quarterly report<|end|>\\

\textbf{\fontsize{11}{13}\selectfont Four-step example}:\\

\textbf{query}: Find and email the project proposal document to project.manager@company.com, then look up the submission deadline and schedule it in my calendar with a reminder.

\textbf{response}: Locate the project proposal document<nexa\_split> Send the project proposal via email to the project manager at project.manager@company.com<nexa\_split> Find the project proposal submission deadline<nexa\_split> Schedule a calendar event for the project proposal submission deadline with a reminder<|end|>\\

\textbf{\fontsize{11}{13}\selectfont Five-step example}:\\

\textbf{query}: I have a meeting tomorrow morning, please find my presentation then connect to the conference room projector via Bluetooth. After that, increase the screen brightness then take a screenshot of the final summary slide and email it to all participants.

\textbf{response}: Find the presentation for the meeting tomorrow<nexa\_split> Connect to the conference room projector via Bluetooth<nexa\_split> Increase screen brightness by 20\%<nexa\_split> Take a screenshot of the final summary slide<nexa\_split> Email the screenshot to all meeting participants<|end|>
\end{tcolorbox}

A visualization of the dataset collection process is shown in Figure \ref{fig:data}. Example function descriptions are provided in Appendix~\ref{sec:a1}.

\begin{figure}[ht]
    \centering
    \includegraphics[width=1.0\textwidth]{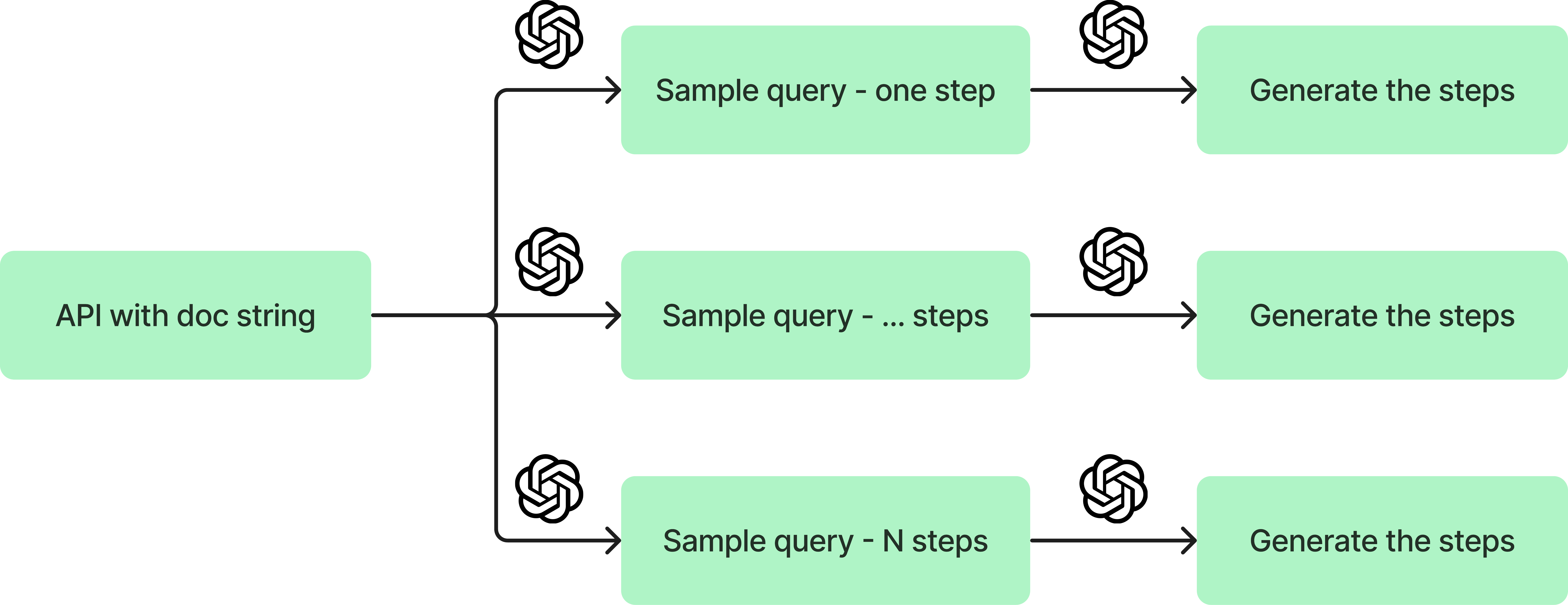}
    \caption{Dataset Collection Process for Planner Model Training. First, we identify the number of steps required, setting $N$ from 1 to 5 in our current case. Next, we generate corresponding queries and steps for each query.}
    \label{fig:data}
\end{figure}

% Let $\Theta$ denote the weights of our model. Post-training, we transform the initial weights $\Theta_0$ to $\Theta$, incorporating the information from all function descriptions $F$ into the model parameters. While full model training yields high performance, it requires retraining the entire model to incorporate new available functions. To address this limitation, we propose a modular "plug and play" approach. In our experimental setup, we train the model to obtain LoRA parameters $\Delta\Theta_i$ for each function set $F_i$. This method allows for the training of multiple LoRAs, facilitating planning across diverse function sets.

% Our experiments revealed an interesting phenomenon: the combination of multiple LoRAs enables the model to support planning across multiple function sets concurrently. We can formalize the process of integrating these LoRAs as:
% \begin{equation}
% \Theta = \Theta_0 + \sum_{i=0}^{n-1}\Delta \Theta_i
% \end{equation}
% where $\Theta_0$ represents the base model weights and $\Delta\Theta_i$ denotes the LoRA parameters for the $i$-th function set.
% We apply this method to available functions available within a mobile application ecosystem. Our results show that the proposed approach achieves high accuracy on the available functions. Additionally, the multiple LoRA approach demonstrates plug-and-play capability when training the planner model, offering significant flexibility and scalability.

\subsection{Benchmark design}
Our evaluation relies on a carefully constructed test dataset. This dataset is designed to represent the complexities of real-world planning, employing a multi-stage approach that integrates automated generation, expert validation, and empirical testing.

The process begins with the automated generation of an initial dataset comprising 1,000 data points using GPT-4. These data points then undergo a rigorous quality assurance process to ensure their integrity and relevance. The quality assessment criteria are as follows:
\begin{itemize}
\item Each step must correspond to an existing function;
\item The sequential order of steps must be correct.
\end{itemize}

To ensure the reliability of our evaluation, we incorporate an additional phase of manual verification. This phase involves selecting a subset of examples for end-to-end model execution, thereby validating the accuracy of our results and providing a comprehensive assessment of our model's performance.

For the evaluation of our proposed planning model, we employ GPT-4 as an oracle to determine the correctness of the generated plans. This choice is based on empirical observations indicating GPT-4's high proficiency in our specific use case. 
%The resulting high-quality benchmark enables us to effectively assess the efficacy of our model in real-world API planning scenarios.

% \begin{tcolorbox}[size=small]
% <|user|> Find the quarterly report for Q3 and print it for the upcoming board meeting.<|end|>\\

% <|assistant|> Locate the quarterly report document for Q3<|nexa\_split|>Print the located quarterly report document<|end|> <|n\_function|> 2<|end|> <|order\_functions|> locate\_file(file\_description)<|nexa\_split|>print\_file(file\_name)<|end|>
% \end{tcolorbox}

\section{Experimental Design}
Our experimental design assesses Octo-planner's performance in on-device AI agent planning. We aim to determine the optimal configuration for deploying efficient, accurate planning models on resource-constrained devices while maintaining adaptability to new domains and functions. Our experiments focus on four key areas:

\begin{enumerate}
\item Performance and efficiency trade-offs between full fine-tuning and LoRA.
\item Multi-LoRA accuracy in handling different function sets simultaneously.
\item Performance comparison across various base models and sizes.
\item The impact of dataset size on accuracy, ranging from 100 to 1000 training examples.
\end{enumerate}

We conduct supervised fine-tuning on our curated dataset, using Phi-3 Mini and a few other alternatives as base models. Training includes both full fine-tuning and LoRA techniques. For all experiments, we set the dataset size to be 800 times the number of available functions and perform fine-tuning on an NVIDIA A100 GPU. We use optimized hyperparameters across both techniques: a learning rate of $5\times 10^{-6}$, a batch size of 4, and a warm-up ratio of 0.2, training for 2 epochs. For LoRA, we set target\_modules to \texttt{all-linear}. %Our preliminary experiments indicate that 2 epochs are sufficient for convergence.

\section{Results}
\subsection{Full fine-tuning vs LoRA}

Table~\ref{tab:1} presents a detailed comparison of the full fine-tuning and LoRA approaches for our planning model. Our experiments reveal significant differences in performance across these methods. Full fine-tuning achieves the highest accuracy at 98.1\%, demonstrating superior performance. In contrast, LoRA's performance depends on the rank size. With rank 64 and alpha 256, LoRA achieves 85.1\% accuracy, while reducing the rank to 16 and alpha to 32 drops accuracy to 72.9\%.
These results highlight the trade-off between model performance and computational efficiency when using LoRA. While full fine-tuning provides better accuracy, LoRA offers a more resource-efficient alternative, with performance varying according to the rank configuration.

\begin{table}[ht]
  \caption{Full fine-tuning vs LoRA benchmark}
  \label{tab:1}
  \centering
  \begin{tabular}{lc}
    \toprule
     Training Configuration & Benchmark Accuracy \\
    \midrule
    Full model training & 98.1\% \\
    LoRA, rank = 64, lora\_alpha = 256, target\_modules \texttt{"all-linear"} & 85.1\% \\
    LoRA, rank = 16, lora\_alpha = 32, target\_modules \texttt{"all-linear"} & 72.9\% \\
    \bottomrule
  \end{tabular}
\end{table}

\subsection{Multi-LoRA training and merging}
While LoRA-based training is effective for specific function sets, real-world applications often require handling new or expanded sets of functions. To address this challenge, we propose merging multiple LoRA weights, each trained on a distinct function subset, into the same base model. This approach creates a composite model that combines knowledge from various function sets, offering a scalable solution for complex, multi-domain queries in resource-constrained environments.

To evaluate this method, we constructed a benchmark dataset by randomly selecting functions from each LoRA's domain with equal probability, combining them into workflows. Queries and plans were generated using GPT-4. For example, when testing two merged LoRAs, queries may involve Android functions, E-commerce functions, or both, with equal likelihood.

The code block below shows sample queries from our benchmark dataset and the corresponding inference results from the multi-LoRA merged model:

\begin{center}
\begin{tcolorbox}
\textbf{\fontsize{11}{13}\selectfont Android functions only}:\\

\textbf{query}:  Find the quarterly report for Q3 and print it for the upcoming board meeting.

\textbf{response}: Locate the quarterly report document for Q3<|nexa\_split|>Print the located quarterly report document<|end|>\\

\textbf{\fontsize{11}{13}\selectfont E-commerce functions only}:\\

\textbf{query}: Find ergonomic office chairs and then buy the selected one using my default shipping address and credit card

\textbf{response}: Search for ergonomic office chairs<nexa\_split>Purchase the selected ergonomic office chair using the default shipping address and credit card<|end|>\\

\textbf{\fontsize{11}{13}\selectfont Android and E-commerce functions}:\\

\textbf{query}: Find ergonomic office chair list doc and print it for my shopping

\textbf{response}: Search for ergonomic office chair list doc<nexa\_split>Print the ergonomic office chair list doc <|end|>
\end{tcolorbox}
\end{center}

Table \ref{tab:2} presents the performance results of our multi-LoRA merging technique. Each individual LoRA was trained with consistent hyperparameters: rank 64, lora\_alpha 256, and target\_modules set to \texttt{"all-linear"}.
The single-domain Android function set LoRA achieves 85.1\% accuracy. When merging LoRAs from two domains (Android and E-commerce), accuracy slightly decreases to 82.2\%. Further merging yields lower accuracies: 78.9\% for three domains (adding Video Streaming) and 69.7\% for four domains (adding Travel). These results reveal a pattern of gradual accuracy decline as more function sets are integrated, with a steeper drop occurring after the third domain is added.

\begin{table}[ht]
  \caption{Multi-LoRA Benchmark}
  \label{tab:2}
  \centering
  \begin{tabular}{lc}
    \toprule
     Training Configuration & Benchmark Accuracy (\%) \\
    \midrule
    LoRA for Android & 85.1 \\
    Merged for Android, E-commerce & 82.2 \\
    Merged for Android, E-commerce, Video Streaming & 78.9 \\
    Merged for Android, E-commerce, Video Streaming, Travel & 69.7 \\
    \bottomrule
  \end{tabular}
\end{table}

\subsection{Full fine-tuning with different base models}

Table \ref{tab:3} presents the benchmark accuracy of different base models after full fine-tuning. Google Gemma 2B achieves 85.6\% accuracy, while the larger Gemma 7B excels with 99.7\%. Microsoft Phi-3 Mini also performs strongly at 98.1\%. These results indicate that our framework adapts well to various on-device LLMs, with larger models generally achieving higher accuracy.

\begin{table}[ht]
  \caption{Different base model benchmark}
  \label{tab:3}
  \centering
  \begin{tabular}{lc}
    \toprule
     Base model & Benchmark Accuracy \\
    \midrule
    Google Gemma 2B & 85.6\% \\
    Google Gemma 7B & 99.7\% \\
    Microsoft Phi-3 Mini & 98.1\%  \\
    \bottomrule
  \end{tabular}
\end{table}

\subsection{Full fine-tuning with different dataset sizes}
Our default training dataset contains 1000 data points, evenly distributed across 1--5 step sequences (200 each) to represent varying task complexities. We investigate the impact of dataset size on model performance to optimize the efficiency of function set integration and address the cost of synthetic data generation. Table \ref{tab:4} shows the benchmark accuracy for various training dataset sizes:

\begin{table}[ht]
  \caption{Different training dataset size benchmark}
  \label{tab:4}
  \centering
  \begin{tabular}{cc}
    \toprule
     Training Dataset Size & Benchmark Accuracy \\
    \midrule
    1000 & 98.1\% \\
    500 & 92.5\% \\
    250 & 85.3\% \\
    100 & 78.1\% \\
    \bottomrule
  \end{tabular}
\end{table}

The results show a clear correlation between dataset size and accuracy. The full 1000-point dataset achieves 98.1\% accuracy, while reducing the dataset to 500 data points drops accuracy to 92.5\%. Further reductions to 250 and 100 data points result in accuracies of 85.3\% and 78.1\%, respectively. These findings suggest that, for optimal performance, a training dataset of more than 1000 data points is recommended.

\section{Conclusion}
This paper introduces Octo-planner, an on-device planning agent designed to work alongside action agents such as Octopus V2. By separating planning and action execution, we improve specialization and adaptability. Our approach fine-tunes Phi-3 Mini (a 3.8-billion-parameter LLM) to serve as a planning agent capable of running locally on edge devices, achieving 97\% success in in-domain tests. We reduce computational demands, improving latency and battery life, and implement a multi-LoRA technique for expanding model capabilities without full retraining.

Octo-planner contributes to addressing AI deployment concerns, including data privacy, latency, and offline functionality. It represents an advancement towards practical, sophisticated AI agents for personal devices. By open-sourcing our model weights, we aim to drive innovation in on-device AI, promoting the development of efficient, privacy-respecting applications that enhance daily life without compromising performance or security.

\section*{Limitations and future work}
Our current model, while effective for specific mobile phone use cases, has limitations in its broader applicability. Unlike frameworks such as ReAct, which alternates between planning steps and executing actions based on real-time feedback, our model conducts all of its planning in advance. This upfront planning approach, while efficient for straightforward tasks, may be less adaptable to complex or unpredictable scenarios in which conditions change during execution.

Future work will focus on exploring an iterative planning methodology that refines plans based on real-time observations, improving adaptability to dynamic environments. We also plan to investigate the integration of our planning model with diverse action models, extending its capabilities beyond mobile applications to areas such as IoT, robotics, and smart home systems. These advancements will address current limitations and expand the versatility of our on-device planning model, bridging the gap between efficient, localized AI processing and the complex demands of real-world applications.

% \newpage
% \medskip
% {\small
\bibliography{citation}

\begin{thebibliography}{10}

\bibitem{jennings1998applications}
Nicholas~R Jennings and Michael Wooldridge.
\newblock Applications of intelligent agents.
\newblock {\em Agent technology: foundations, applications, and markets}, pages
  3--28, 1998.

\bibitem{poole2010artificial}
David~L Poole and Alan~K Mackworth.
\newblock {\em Artificial Intelligence: foundations of computational agents}.
\newblock Cambridge University Press, 2010.

\bibitem{kim2023language}
Geunwoo Kim, Pierre Baldi, and Stephen McAleer.
\newblock Language models can solve computer tasks, 2023.

\bibitem{deng2023mind2web}
Xiang Deng, Yu~Gu, Boyuan Zheng, Shijie Chen, Samuel Stevens, Boshi Wang, Huan
  Sun, and Yu~Su.
\newblock Mind2web: Towards a generalist agent for the web, 2023.

\bibitem{yan2023gpt4v}
An~Yan, Zhengyuan Yang, Wanrong Zhu, Kevin Lin, Linjie Li, Jianfeng Wang,
  Jianwei Yang, Yiwu Zhong, Julian McAuley, Jianfeng Gao, Zicheng Liu, and
  Lijuan Wang.
\newblock Gpt-4v in wonderland: Large multimodal models for zero-shot
  smartphone gui navigation, 2023.

\bibitem{zheng2024gpt4vision}
Boyuan Zheng, Boyu Gou, Jihyung Kil, Huan Sun, and Yu~Su.
\newblock Gpt-4v(ision) is a generalist web agent, if grounded, 2024.

\bibitem{koh2024visualwebarena}
Jing~Yu Koh, Robert Lo, Lawrence Jang, Vikram Duvvur, Ming~Chong Lim, Po-Yu
  Huang, Graham Neubig, Shuyan Zhou, Ruslan Salakhutdinov, and Daniel Fried.
\newblock Visualwebarena: Evaluating multimodal agents on realistic visual web
  tasks, 2024.

\bibitem{geminiteam2024gemini}
Gemini Team, Rohan Anil, Sebastian Borgeaud, Jean-Baptiste Alayrac, Jiahui Yu,
  Radu Soricut, Johan Schalkwyk, Andrew~M. Dai, Anja Hauth, Katie Millican,
  David Silver, Melvin Johnson, Ioannis Antonoglou, Julian Schrittwieser,
  Amelia Glaese, Jilin Chen, Emily Pitler, Timothy Lillicrap, Angeliki
  Lazaridou, Orhan Firat, James Molloy, Michael Isard, Paul~R. Barham, Tom
  Hennigan, Benjamin Lee, Fabio Viola, Malcolm Reynolds, Yuanzhong Xu, Ryan
  Doherty, Eli Collins, Clemens Meyer, Eliza Rutherford, Erica Moreira, Kareem
  Ayoub, Megha Goel, Jack Krawczyk, Cosmo Du, Ed~Chi, Heng-Tze Cheng, Eric Ni,
  Purvi Shah, Patrick Kane, Betty Chan, Manaal Faruqui, Aliaksei Severyn,
  Hanzhao Lin, YaGuang Li, Yong Cheng, Abe Ittycheriah, Mahdis Mahdieh, Mia
  Chen, Pei Sun, Dustin Tran, Sumit Bagri, Balaji Lakshminarayanan, Jeremiah
  Liu, Andras Orban, Fabian Güra, Hao Zhou, Xinying Song, Aurelien Boffy,
  Harish Ganapathy, Steven Zheng, HyunJeong Choe, Ágoston Weisz, Tao Zhu,
  Yifeng Lu, Siddharth Gopal, Jarrod Kahn, Maciej Kula, Jeff Pitman, Rushin
  Shah, Emanuel Taropa, Majd~Al Merey, Martin Baeuml, Zhifeng Chen, Laurent~El
  Shafey, Yujing Zhang, Olcan Sercinoglu, George Tucker, Enrique Piqueras,
  Maxim Krikun, Iain Barr, Nikolay Savinov, Ivo Danihelka, Becca Roelofs,
  Anaïs White, Anders Andreassen, Tamara von Glehn, Lakshman Yagati, Mehran
  Kazemi, Lucas Gonzalez, Misha Khalman, Jakub Sygnowski, Alexandre Frechette,
  Charlotte Smith, Laura Culp, Lev Proleev, Yi~Luan, Xi~Chen, James Lottes,
  Nathan Schucher, Federico Lebron, Alban Rrustemi, Natalie Clay, Phil Crone,
  Tomas Kocisky, Jeffrey Zhao, Bartek Perz, Dian Yu, Heidi Howard, Adam
  Bloniarz, Jack~W. Rae, Han Lu, Laurent Sifre, Marcello Maggioni, Fred
  Alcober, Dan Garrette, Megan Barnes, Shantanu Thakoor, Jacob Austin, Gabriel
  Barth-Maron, William Wong, Rishabh Joshi, Rahma Chaabouni, Deeni Fatiha, Arun
  Ahuja, Gaurav~Singh Tomar, Evan Senter, Martin Chadwick, Ilya Kornakov,
  Nithya Attaluri, Iñaki Iturrate, Ruibo Liu, Yunxuan Li, Sarah Cogan, Jeremy
  Chen, Chao Jia, Chenjie Gu, Qiao Zhang, Jordan Grimstad, Ale~Jakse Hartman,
  Xavier Garcia, Thanumalayan~Sankaranarayana Pillai, Jacob Devlin, Michael
  Laskin, Diego de~Las~Casas, Dasha Valter, Connie Tao, Lorenzo Blanco,
  Adrià~Puigdomènech Badia, David Reitter, Mianna Chen, Jenny Brennan, Clara
  Rivera, Sergey Brin, Shariq Iqbal, Gabriela Surita, Jane Labanowski, Abhi
  Rao, Stephanie Winkler, Emilio Parisotto, Yiming Gu, Kate Olszewska, Ravi
  Addanki, Antoine Miech, Annie Louis, Denis Teplyashin, Geoff Brown, Elliot
  Catt, Jan Balaguer, Jackie Xiang, Pidong Wang, Zoe Ashwood, Anton Briukhov,
  Albert Webson, Sanjay Ganapathy, Smit Sanghavi, Ajay Kannan, Ming-Wei Chang,
  Axel Stjerngren, Josip Djolonga, Yuting Sun, Ankur Bapna, Matthew Aitchison,
  Pedram Pejman, Henryk Michalewski, Tianhe Yu, Cindy Wang, Juliette Love,
  Junwhan Ahn, Dawn Bloxwich, Kehang Han, Peter Humphreys, Thibault Sellam,
  James Bradbury, Varun Godbole, Sina Samangooei, Bogdan Damoc, Alex Kaskasoli,
  Sébastien M.~R. Arnold, Vijay Vasudevan, Shubham Agrawal, Jason Riesa,
  Dmitry Lepikhin, Richard Tanburn, Srivatsan Srinivasan, Hyeontaek Lim, Sarah
  Hodkinson, Pranav Shyam, Johan Ferret, Steven Hand, Ankush Garg, Tom~Le
  Paine, Jian Li, Yujia Li, Minh Giang, Alexander Neitz, Zaheer Abbas, Sarah
  York, Machel Reid, Elizabeth Cole, Aakanksha Chowdhery, Dipanjan Das,
  Dominika Rogozińska, Vitaliy Nikolaev, Pablo Sprechmann, Zachary Nado, Lukas
  Zilka, Flavien Prost, Luheng He, Marianne Monteiro, Gaurav Mishra, Chris
  Welty, Josh Newlan, Dawei Jia, Miltiadis Allamanis, Clara~Huiyi Hu, Raoul
  de~Liedekerke, Justin Gilmer, Carl Saroufim, Shruti Rijhwani, Shaobo Hou,
  Disha Shrivastava, Anirudh Baddepudi, Alex Goldin, Adnan Ozturel, Albin
  Cassirer, Yunhan Xu, Daniel Sohn, Devendra Sachan, Reinald~Kim Amplayo, Craig
  Swanson, Dessie Petrova, Shashi Narayan, Arthur Guez, Siddhartha Brahma,
  Jessica Landon, Miteyan Patel, Ruizhe Zhao, Kevin Villela, Luyu Wang, Wenhao
  Jia, Matthew Rahtz, Mai Giménez, Legg Yeung, James Keeling, Petko Georgiev,
  Diana Mincu, Boxi Wu, Salem Haykal, Rachel Saputro, Kiran Vodrahalli, James
  Qin, Zeynep Cankara, Abhanshu Sharma, Nick Fernando, Will Hawkins, Behnam
  Neyshabur, Solomon Kim, Adrian Hutter, Priyanka Agrawal, Alex Castro-Ros,
  George van~den Driessche, Tao Wang, Fan Yang, Shuo yiin Chang, Paul Komarek,
  Ross McIlroy, Mario Lučić, Guodong Zhang, Wael Farhan, Michael Sharman,
  Paul Natsev, Paul Michel, Yamini Bansal, Siyuan Qiao, Kris Cao, Siamak
  Shakeri, Christina Butterfield, Justin Chung, Paul~Kishan Rubenstein, Shivani
  Agrawal, Arthur Mensch, Kedar Soparkar, Karel Lenc, Timothy Chung, Aedan
  Pope, Loren Maggiore, Jackie Kay, Priya Jhakra, Shibo Wang, Joshua Maynez,
  Mary Phuong, Taylor Tobin, Andrea Tacchetti, Maja Trebacz, Kevin Robinson,
  Yash Katariya, Sebastian Riedel, Paige Bailey, Kefan Xiao, Nimesh Ghelani,
  Lora Aroyo, Ambrose Slone, Neil Houlsby, Xuehan Xiong, Zhen Yang, Elena
  Gribovskaya, Jonas Adler, Mateo Wirth, Lisa Lee, Music Li, Thais Kagohara,
  Jay Pavagadhi, Sophie Bridgers, Anna Bortsova, Sanjay Ghemawat, Zafarali
  Ahmed, Tianqi Liu, Richard Powell, Vijay Bolina, Mariko Iinuma, Polina
  Zablotskaia, James Besley, Da-Woon Chung, Timothy Dozat, Ramona Comanescu,
  Xiance Si, Jeremy Greer, Guolong Su, Martin Polacek, Raphaël~Lopez Kaufman,
  Simon Tokumine, Hexiang Hu, Elena Buchatskaya, Yingjie Miao, Mohamed
  Elhawaty, Aditya Siddhant, Nenad Tomasev, Jinwei Xing, Christina Greer, Helen
  Miller, Shereen Ashraf, Aurko Roy, Zizhao Zhang, Ada Ma, Angelos Filos, Milos
  Besta, Rory Blevins, Ted Klimenko, Chih-Kuan Yeh, Soravit Changpinyo, Jiaqi
  Mu, Oscar Chang, Mantas Pajarskas, Carrie Muir, Vered Cohen, Charline~Le Lan,
  Krishna Haridasan, Amit Marathe, Steven Hansen, Sholto Douglas, Rajkumar
  Samuel, Mingqiu Wang, Sophia Austin, Chang Lan, Jiepu Jiang, Justin Chiu,
  Jaime~Alonso Lorenzo, Lars~Lowe Sjösund, Sébastien Cevey, Zach Gleicher,
  Thi Avrahami, Anudhyan Boral, Hansa Srinivasan, Vittorio Selo, Rhys May,
  Konstantinos Aisopos, Léonard Hussenot, Livio~Baldini Soares, Kate Baumli,
  Michael~B. Chang, Adrià Recasens, Ben Caine, Alexander Pritzel, Filip
  Pavetic, Fabio Pardo, Anita Gergely, Justin Frye, Vinay Ramasesh, Dan Horgan,
  Kartikeya Badola, Nora Kassner, Subhrajit Roy, Ethan Dyer, Víctor~Campos
  Campos, Alex Tomala, Yunhao Tang, Dalia~El Badawy, Elspeth White, Basil
  Mustafa, Oran Lang, Abhishek Jindal, Sharad Vikram, Zhitao Gong, Sergi
  Caelles, Ross Hemsley, Gregory Thornton, Fangxiaoyu Feng, Wojciech Stokowiec,
  Ce~Zheng, Phoebe Thacker, Çağlar Ünlü, Zhishuai Zhang, Mohammad Saleh,
  James Svensson, Max Bileschi, Piyush Patil, Ankesh Anand, Roman Ring,
  Katerina Tsihlas, Arpi Vezer, Marco Selvi, Toby Shevlane, Mikel Rodriguez,
  Tom Kwiatkowski, Samira Daruki, Keran Rong, Allan Dafoe, Nicholas FitzGerald,
  Keren Gu-Lemberg, Mina Khan, Lisa~Anne Hendricks, Marie Pellat, Vladimir
  Feinberg, James Cobon-Kerr, Tara Sainath, Maribeth Rauh, Sayed~Hadi Hashemi,
  Richard Ives, Yana Hasson, Eric Noland, Yuan Cao, Nathan Byrd, Le~Hou, Qingze
  Wang, Thibault Sottiaux, Michela Paganini, Jean-Baptiste Lespiau, Alexandre
  Moufarek, Samer Hassan, Kaushik Shivakumar, Joost van Amersfoort, Amol
  Mandhane, Pratik Joshi, Anirudh Goyal, Matthew Tung, Andrew Brock, Hannah
  Sheahan, Vedant Misra, Cheng Li, Nemanja Rakićević, Mostafa Dehghani,
  Fangyu Liu, Sid Mittal, Junhyuk Oh, Seb Noury, Eren Sezener, Fantine Huot,
  Matthew Lamm, Nicola~De Cao, Charlie Chen, Sidharth Mudgal, Romina Stella,
  Kevin Brooks, Gautam Vasudevan, Chenxi Liu, Mainak Chain, Nivedita Melinkeri,
  Aaron Cohen, Venus Wang, Kristie Seymore, Sergey Zubkov, Rahul Goel, Summer
  Yue, Sai Krishnakumaran, Brian Albert, Nate Hurley, Motoki Sano, Anhad
  Mohananey, Jonah Joughin, Egor Filonov, Tomasz Kępa, Yomna Eldawy, Jiawern
  Lim, Rahul Rishi, Shirin Badiezadegan, Taylor Bos, Jerry Chang, Sanil Jain,
  Sri Gayatri~Sundara Padmanabhan, Subha Puttagunta, Kalpesh Krishna, Leslie
  Baker, Norbert Kalb, Vamsi Bedapudi, Adam Kurzrok, Shuntong Lei, Anthony Yu,
  Oren Litvin, Xiang Zhou, Zhichun Wu, Sam Sobell, Andrea Siciliano, Alan
  Papir, Robby Neale, Jonas Bragagnolo, Tej Toor, Tina Chen, Valentin Anklin,
  Feiran Wang, Richie Feng, Milad Gholami, Kevin Ling, Lijuan Liu, Jules
  Walter, Hamid Moghaddam, Arun Kishore, Jakub Adamek, Tyler Mercado, Jonathan
  Mallinson, Siddhinita Wandekar, Stephen Cagle, Eran Ofek, Guillermo Garrido,
  Clemens Lombriser, Maksim Mukha, Botu Sun, Hafeezul~Rahman Mohammad, Josip
  Matak, Yadi Qian, Vikas Peswani, Pawel Janus, Quan Yuan, Leif Schelin, Oana
  David, Ankur Garg, Yifan He, Oleksii Duzhyi, Anton Älgmyr, Timothée Lottaz,
  Qi~Li, Vikas Yadav, Luyao Xu, Alex Chinien, Rakesh Shivanna, Aleksandr
  Chuklin, Josie Li, Carrie Spadine, Travis Wolfe, Kareem Mohamed, Subhabrata
  Das, Zihang Dai, Kyle He, Daniel von Dincklage, Shyam Upadhyay, Akanksha
  Maurya, Luyan Chi, Sebastian Krause, Khalid Salama, Pam~G Rabinovitch, Pavan
  Kumar~Reddy M, Aarush Selvan, Mikhail Dektiarev, Golnaz Ghiasi, Erdem Guven,
  Himanshu Gupta, Boyi Liu, Deepak Sharma, Idan~Heimlich Shtacher, Shachi Paul,
  Oscar Akerlund, François-Xavier Aubet, Terry Huang, Chen Zhu, Eric Zhu,
  Elico Teixeira, Matthew Fritze, Francesco Bertolini, Liana-Eleonora
  Marinescu, Martin Bölle, Dominik Paulus, Khyatti Gupta, Tejasi Latkar, Max
  Chang, Jason Sanders, Roopa Wilson, Xuewei Wu, Yi-Xuan Tan, Lam~Nguyen Thiet,
  Tulsee Doshi, Sid Lall, Swaroop Mishra, Wanming Chen, Thang Luong, Seth
  Benjamin, Jasmine Lee, Ewa Andrejczuk, Dominik Rabiej, Vipul Ranjan,
  Krzysztof Styrc, Pengcheng Yin, Jon Simon, Malcolm~Rose Harriott, Mudit
  Bansal, Alexei Robsky, Geoff Bacon, David Greene, Daniil Mirylenka, Chen
  Zhou, Obaid Sarvana, Abhimanyu Goyal, Samuel Andermatt, Patrick Siegler, Ben
  Horn, Assaf Israel, Francesco Pongetti, Chih-Wei~"Louis" Chen, Marco
  Selvatici, Pedro Silva, Kathie Wang, Jackson Tolins, Kelvin Guu, Roey Yogev,
  Xiaochen Cai, Alessandro Agostini, Maulik Shah, Hung Nguyen, Noah~Ó
  Donnaile, Sébastien Pereira, Linda Friso, Adam Stambler, Adam Kurzrok,
  Chenkai Kuang, Yan Romanikhin, Mark Geller, ZJ~Yan, Kane Jang, Cheng-Chun
  Lee, Wojciech Fica, Eric Malmi, Qijun Tan, Dan Banica, Daniel Balle, Ryan
  Pham, Yanping Huang, Diana Avram, Hongzhi Shi, Jasjot Singh, Chris Hidey,
  Niharika Ahuja, Pranab Saxena, Dan Dooley, Srividya~Pranavi Potharaju, Eileen
  O'Neill, Anand Gokulchandran, Ryan Foley, Kai Zhao, Mike Dusenberry, Yuan
  Liu, Pulkit Mehta, Ragha Kotikalapudi, Chalence Safranek-Shrader, Andrew
  Goodman, Joshua Kessinger, Eran Globen, Prateek Kolhar, Chris Gorgolewski,
  Ali Ibrahim, Yang Song, Ali Eichenbaum, Thomas Brovelli, Sahitya Potluri,
  Preethi Lahoti, Cip Baetu, Ali Ghorbani, Charles Chen, Andy Crawford, Shalini
  Pal, Mukund Sridhar, Petru Gurita, Asier Mujika, Igor Petrovski, Pierre-Louis
  Cedoz, Chenmei Li, Shiyuan Chen, Niccolò~Dal Santo, Siddharth Goyal, Jitesh
  Punjabi, Karthik Kappaganthu, Chester Kwak, Pallavi LV, Sarmishta Velury,
  Himadri Choudhury, Jamie Hall, Premal Shah, Ricardo Figueira, Matt Thomas,
  Minjie Lu, Ting Zhou, Chintu Kumar, Thomas Jurdi, Sharat Chikkerur, Yenai Ma,
  Adams Yu, Soo Kwak, Victor Ähdel, Sujeevan Rajayogam, Travis Choma, Fei Liu,
  Aditya Barua, Colin Ji, Ji~Ho Park, Vincent Hellendoorn, Alex Bailey, Taylan
  Bilal, Huanjie Zhou, Mehrdad Khatir, Charles Sutton, Wojciech Rzadkowski,
  Fiona Macintosh, Konstantin Shagin, Paul Medina, Chen Liang, Jinjing Zhou,
  Pararth Shah, Yingying Bi, Attila Dankovics, Shipra Banga, Sabine Lehmann,
  Marissa Bredesen, Zifan Lin, John~Eric Hoffmann, Jonathan Lai, Raynald Chung,
  Kai Yang, Nihal Balani, Arthur Bražinskas, Andrei Sozanschi, Matthew Hayes,
  Héctor~Fernández Alcalde, Peter Makarov, Will Chen, Antonio Stella,
  Liselotte Snijders, Michael Mandl, Ante Kärrman, Paweł Nowak, Xinyi Wu,
  Alex Dyck, Krishnan Vaidyanathan, Raghavender R, Jessica Mallet, Mitch
  Rudominer, Eric Johnston, Sushil Mittal, Akhil Udathu, Janara Christensen,
  Vishal Verma, Zach Irving, Andreas Santucci, Gamaleldin Elsayed, Elnaz
  Davoodi, Marin Georgiev, Ian Tenney, Nan Hua, Geoffrey Cideron, Edouard
  Leurent, Mahmoud Alnahlawi, Ionut Georgescu, Nan Wei, Ivy Zheng, Dylan
  Scandinaro, Heinrich Jiang, Jasper Snoek, Mukund Sundararajan, Xuezhi Wang,
  Zack Ontiveros, Itay Karo, Jeremy Cole, Vinu Rajashekhar, Lara Tumeh, Eyal
  Ben-David, Rishub Jain, Jonathan Uesato, Romina Datta, Oskar Bunyan, Shimu
  Wu, John Zhang, Piotr Stanczyk, Ye~Zhang, David Steiner, Subhajit Naskar,
  Michael Azzam, Matthew Johnson, Adam Paszke, Chung-Cheng Chiu, Jaume~Sanchez
  Elias, Afroz Mohiuddin, Faizan Muhammad, Jin Miao, Andrew Lee, Nino
  Vieillard, Jane Park, Jiageng Zhang, Jeff Stanway, Drew Garmon, Abhijit
  Karmarkar, Zhe Dong, Jong Lee, Aviral Kumar, Luowei Zhou, Jonathan Evens,
  William Isaac, Geoffrey Irving, Edward Loper, Michael Fink, Isha Arkatkar,
  Nanxin Chen, Izhak Shafran, Ivan Petrychenko, Zhe Chen, Johnson Jia, Anselm
  Levskaya, Zhenkai Zhu, Peter Grabowski, Yu~Mao, Alberto Magni, Kaisheng Yao,
  Javier Snaider, Norman Casagrande, Evan Palmer, Paul Suganthan, Alfonso
  Castaño, Irene Giannoumis, Wooyeol Kim, Mikołaj Rybiński, Ashwin
  Sreevatsa, Jennifer Prendki, David Soergel, Adrian Goedeckemeyer, Willi
  Gierke, Mohsen Jafari, Meenu Gaba, Jeremy Wiesner, Diana~Gage Wright, Yawen
  Wei, Harsha Vashisht, Yana Kulizhskaya, Jay Hoover, Maigo Le, Lu~Li, Chimezie
  Iwuanyanwu, Lu~Liu, Kevin Ramirez, Andrey Khorlin, Albert Cui, Tian LIN,
  Marcus Wu, Ricardo Aguilar, Keith Pallo, Abhishek Chakladar, Ginger Perng,
  Elena~Allica Abellan, Mingyang Zhang, Ishita Dasgupta, Nate Kushman, Ivo
  Penchev, Alena Repina, Xihui Wu, Tom van~der Weide, Priya Ponnapalli,
  Caroline Kaplan, Jiri Simsa, Shuangfeng Li, Olivier Dousse, Fan Yang, Jeff
  Piper, Nathan Ie, Rama Pasumarthi, Nathan Lintz, Anitha Vijayakumar, Daniel
  Andor, Pedro Valenzuela, Minnie Lui, Cosmin Paduraru, Daiyi Peng, Katherine
  Lee, Shuyuan Zhang, Somer Greene, Duc~Dung Nguyen, Paula Kurylowicz, Cassidy
  Hardin, Lucas Dixon, Lili Janzer, Kiam Choo, Ziqiang Feng, Biao Zhang,
  Achintya Singhal, Dayou Du, Dan McKinnon, Natasha Antropova, Tolga Bolukbasi,
  Orgad Keller, David Reid, Daniel Finchelstein, Maria~Abi Raad, Remi Crocker,
  Peter Hawkins, Robert Dadashi, Colin Gaffney, Ken Franko, Anna Bulanova,
  Rémi Leblond, Shirley Chung, Harry Askham, Luis~C. Cobo, Kelvin Xu, Felix
  Fischer, Jun Xu, Christina Sorokin, Chris Alberti, Chu-Cheng Lin, Colin
  Evans, Alek Dimitriev, Hannah Forbes, Dylan Banarse, Zora Tung, Mark
  Omernick, Colton Bishop, Rachel Sterneck, Rohan Jain, Jiawei Xia, Ehsan Amid,
  Francesco Piccinno, Xingyu Wang, Praseem Banzal, Daniel~J. Mankowitz, Alex
  Polozov, Victoria Krakovna, Sasha Brown, MohammadHossein Bateni, Dennis Duan,
  Vlad Firoiu, Meghana Thotakuri, Tom Natan, Matthieu Geist, Ser tan Girgin,
  Hui Li, Jiayu Ye, Ofir Roval, Reiko Tojo, Michael Kwong, James Lee-Thorp,
  Christopher Yew, Danila Sinopalnikov, Sabela Ramos, John Mellor, Abhishek
  Sharma, Kathy Wu, David Miller, Nicolas Sonnerat, Denis Vnukov, Rory Greig,
  Jennifer Beattie, Emily Caveness, Libin Bai, Julian Eisenschlos, Alex
  Korchemniy, Tomy Tsai, Mimi Jasarevic, Weize Kong, Phuong Dao, Zeyu Zheng,
  Frederick Liu, Fan Yang, Rui Zhu, Tian~Huey Teh, Jason Sanmiya, Evgeny
  Gladchenko, Nejc Trdin, Daniel Toyama, Evan Rosen, Sasan Tavakkol, Linting
  Xue, Chen Elkind, Oliver Woodman, John Carpenter, George Papamakarios, Rupert
  Kemp, Sushant Kafle, Tanya Grunina, Rishika Sinha, Alice Talbert, Diane Wu,
  Denese Owusu-Afriyie, Cosmo Du, Chloe Thornton, Jordi Pont-Tuset, Pradyumna
  Narayana, Jing Li, Saaber Fatehi, John Wieting, Omar Ajmeri, Benigno Uria,
  Yeongil Ko, Laura Knight, Amélie Héliou, Ning Niu, Shane Gu, Chenxi Pang,
  Yeqing Li, Nir Levine, Ariel Stolovich, Rebeca Santamaria-Fernandez, Sonam
  Goenka, Wenny Yustalim, Robin Strudel, Ali Elqursh, Charlie Deck, Hyo Lee,
  Zonglin Li, Kyle Levin, Raphael Hoffmann, Dan Holtmann-Rice, Olivier Bachem,
  Sho Arora, Christy Koh, Soheil~Hassas Yeganeh, Siim Põder, Mukarram Tariq,
  Yanhua Sun, Lucian Ionita, Mojtaba Seyedhosseini, Pouya Tafti, Zhiyu Liu,
  Anmol Gulati, Jasmine Liu, Xinyu Ye, Bart Chrzaszcz, Lily Wang, Nikhil Sethi,
  Tianrun Li, Ben Brown, Shreya Singh, Wei Fan, Aaron Parisi, Joe Stanton,
  Vinod Koverkathu, Christopher~A. Choquette-Choo, Yunjie Li, TJ~Lu, Abe
  Ittycheriah, Prakash Shroff, Mani Varadarajan, Sanaz Bahargam, Rob
  Willoughby, David Gaddy, Guillaume Desjardins, Marco Cornero, Brona Robenek,
  Bhavishya Mittal, Ben Albrecht, Ashish Shenoy, Fedor Moiseev, Henrik
  Jacobsson, Alireza Ghaffarkhah, Morgane Rivière, Alanna Walton, Clément
  Crepy, Alicia Parrish, Zongwei Zhou, Clement Farabet, Carey Radebaugh,
  Praveen Srinivasan, Claudia van~der Salm, Andreas Fidjeland, Salvatore
  Scellato, Eri Latorre-Chimoto, Hanna Klimczak-Plucińska, David Bridson,
  Dario de~Cesare, Tom Hudson, Piermaria Mendolicchio, Lexi Walker, Alex
  Morris, Matthew Mauger, Alexey Guseynov, Alison Reid, Seth Odoom, Lucia
  Loher, Victor Cotruta, Madhavi Yenugula, Dominik Grewe, Anastasia
  Petrushkina, Tom Duerig, Antonio Sanchez, Steve Yadlowsky, Amy Shen, Amir
  Globerson, Lynette Webb, Sahil Dua, Dong Li, Surya Bhupatiraju, Dan Hurt,
  Haroon Qureshi, Ananth Agarwal, Tomer Shani, Matan Eyal, Anuj Khare,
  Shreyas~Rammohan Belle, Lei Wang, Chetan Tekur, Mihir~Sanjay Kale, Jinliang
  Wei, Ruoxin Sang, Brennan Saeta, Tyler Liechty, Yi~Sun, Yao Zhao, Stephan
  Lee, Pandu Nayak, Doug Fritz, Manish~Reddy Vuyyuru, John Aslanides, Nidhi
  Vyas, Martin Wicke, Xiao Ma, Evgenii Eltyshev, Nina Martin, Hardie Cate,
  James Manyika, Keyvan Amiri, Yelin Kim, Xi~Xiong, Kai Kang, Florian Luisier,
  Nilesh Tripuraneni, David Madras, Mandy Guo, Austin Waters, Oliver Wang,
  Joshua Ainslie, Jason Baldridge, Han Zhang, Garima Pruthi, Jakob Bauer, Feng
  Yang, Riham Mansour, Jason Gelman, Yang Xu, George Polovets, Ji~Liu, Honglong
  Cai, Warren Chen, XiangHai Sheng, Emily Xue, Sherjil Ozair, Christof
  Angermueller, Xiaowei Li, Anoop Sinha, Weiren Wang, Julia Wiesinger,
  Emmanouil Koukoumidis, Yuan Tian, Anand Iyer, Madhu Gurumurthy, Mark
  Goldenson, Parashar Shah, MK~Blake, Hongkun Yu, Anthony Urbanowicz,
  Jennimaria Palomaki, Chrisantha Fernando, Ken Durden, Harsh Mehta, Nikola
  Momchev, Elahe Rahimtoroghi, Maria Georgaki, Amit Raul, Sebastian Ruder,
  Morgan Redshaw, Jinhyuk Lee, Denny Zhou, Komal Jalan, Dinghua Li, Blake
  Hechtman, Parker Schuh, Milad Nasr, Kieran Milan, Vladimir Mikulik, Juliana
  Franco, Tim Green, Nam Nguyen, Joe Kelley, Aroma Mahendru, Andrea Hu, Joshua
  Howland, Ben Vargas, Jeffrey Hui, Kshitij Bansal, Vikram Rao, Rakesh Ghiya,
  Emma Wang, Ke~Ye, Jean~Michel Sarr, Melanie~Moranski Preston, Madeleine
  Elish, Steve Li, Aakash Kaku, Jigar Gupta, Ice Pasupat, Da-Cheng Juan, Milan
  Someswar, Tejvi M., Xinyun Chen, Aida Amini, Alex Fabrikant, Eric Chu, Xuanyi
  Dong, Amruta Muthal, Senaka Buthpitiya, Sarthak Jauhari, Nan Hua, Urvashi
  Khandelwal, Ayal Hitron, Jie Ren, Larissa Rinaldi, Shahar Drath, Avigail
  Dabush, Nan-Jiang Jiang, Harshal Godhia, Uli Sachs, Anthony Chen, Yicheng
  Fan, Hagai Taitelbaum, Hila Noga, Zhuyun Dai, James Wang, Chen Liang, Jenny
  Hamer, Chun-Sung Ferng, Chenel Elkind, Aviel Atias, Paulina Lee, Vít
  Listík, Mathias Carlen, Jan van~de Kerkhof, Marcin Pikus, Krunoslav Zaher,
  Paul Müller, Sasha Zykova, Richard Stefanec, Vitaly Gatsko, Christoph
  Hirnschall, Ashwin Sethi, Xingyu~Federico Xu, Chetan Ahuja, Beth Tsai, Anca
  Stefanoiu, Bo~Feng, Keshav Dhandhania, Manish Katyal, Akshay Gupta, Atharva
  Parulekar, Divya Pitta, Jing Zhao, Vivaan Bhatia, Yashodha Bhavnani, Omar
  Alhadlaq, Xiaolin Li, Peter Danenberg, Dennis Tu, Alex Pine, Vera Filippova,
  Abhipso Ghosh, Ben Limonchik, Bhargava Urala, Chaitanya~Krishna Lanka, Derik
  Clive, Yi~Sun, Edward Li, Hao Wu, Kevin Hongtongsak, Ianna Li, Kalind
  Thakkar, Kuanysh Omarov, Kushal Majmundar, Michael Alverson, Michael
  Kucharski, Mohak Patel, Mudit Jain, Maksim Zabelin, Paolo Pelagatti, Rohan
  Kohli, Saurabh Kumar, Joseph Kim, Swetha Sankar, Vineet Shah, Lakshmi
  Ramachandruni, Xiangkai Zeng, Ben Bariach, Laura Weidinger, Tu~Vu, Amar
  Subramanya, Sissie Hsiao, Demis Hassabis, Koray Kavukcuoglu, Adam Sadovsky,
  Quoc Le, Trevor Strohman, Yonghui Wu, Slav Petrov, Jeffrey Dean, and Oriol
  Vinyals.
\newblock Gemini: A family of highly capable multimodal models, 2024.

\bibitem{openai2024gpt4}
OpenAI, Josh Achiam, Steven Adler, Sandhini Agarwal, Lama Ahmad, Ilge Akkaya,
  Florencia~Leoni Aleman, Diogo Almeida, Janko Altenschmidt, Sam Altman,
  Shyamal Anadkat, Red Avila, Igor Babuschkin, Suchir Balaji, Valerie Balcom,
  Paul Baltescu, Haiming Bao, Mohammad Bavarian, Jeff Belgum, Irwan Bello, Jake
  Berdine, Gabriel Bernadett-Shapiro, Christopher Berner, Lenny Bogdonoff, Oleg
  Boiko, Madelaine Boyd, Anna-Luisa Brakman, Greg Brockman, Tim Brooks, Miles
  Brundage, Kevin Button, Trevor Cai, Rosie Campbell, Andrew Cann, Brittany
  Carey, Chelsea Carlson, Rory Carmichael, Brooke Chan, Che Chang, Fotis
  Chantzis, Derek Chen, Sully Chen, Ruby Chen, Jason Chen, Mark Chen, Ben
  Chess, Chester Cho, Casey Chu, Hyung~Won Chung, Dave Cummings, Jeremiah
  Currier, Yunxing Dai, Cory Decareaux, Thomas Degry, Noah Deutsch, Damien
  Deville, Arka Dhar, David Dohan, Steve Dowling, Sheila Dunning, Adrien
  Ecoffet, Atty Eleti, Tyna Eloundou, David Farhi, Liam Fedus, Niko Felix,
  Simón~Posada Fishman, Juston Forte, Isabella Fulford, Leo Gao, Elie Georges,
  Christian Gibson, Vik Goel, Tarun Gogineni, Gabriel Goh, Rapha Gontijo-Lopes,
  Jonathan Gordon, Morgan Grafstein, Scott Gray, Ryan Greene, Joshua Gross,
  Shixiang~Shane Gu, Yufei Guo, Chris Hallacy, Jesse Han, Jeff Harris, Yuchen
  He, Mike Heaton, Johannes Heidecke, Chris Hesse, Alan Hickey, Wade Hickey,
  Peter Hoeschele, Brandon Houghton, Kenny Hsu, Shengli Hu, Xin Hu, Joost
  Huizinga, Shantanu Jain, Shawn Jain, Joanne Jang, Angela Jiang, Roger Jiang,
  Haozhun Jin, Denny Jin, Shino Jomoto, Billie Jonn, Heewoo Jun, Tomer Kaftan,
  Łukasz Kaiser, Ali Kamali, Ingmar Kanitscheider, Nitish~Shirish Keskar,
  Tabarak Khan, Logan Kilpatrick, Jong~Wook Kim, Christina Kim, Yongjik Kim,
  Jan~Hendrik Kirchner, Jamie Kiros, Matt Knight, Daniel Kokotajlo, Łukasz
  Kondraciuk, Andrew Kondrich, Aris Konstantinidis, Kyle Kosic, Gretchen
  Krueger, Vishal Kuo, Michael Lampe, Ikai Lan, Teddy Lee, Jan Leike, Jade
  Leung, Daniel Levy, Chak~Ming Li, Rachel Lim, Molly Lin, Stephanie Lin,
  Mateusz Litwin, Theresa Lopez, Ryan Lowe, Patricia Lue, Anna Makanju, Kim
  Malfacini, Sam Manning, Todor Markov, Yaniv Markovski, Bianca Martin, Katie
  Mayer, Andrew Mayne, Bob McGrew, Scott~Mayer McKinney, Christine McLeavey,
  Paul McMillan, Jake McNeil, David Medina, Aalok Mehta, Jacob Menick, Luke
  Metz, Andrey Mishchenko, Pamela Mishkin, Vinnie Monaco, Evan Morikawa, Daniel
  Mossing, Tong Mu, Mira Murati, Oleg Murk, David Mély, Ashvin Nair, Reiichiro
  Nakano, Rajeev Nayak, Arvind Neelakantan, Richard Ngo, Hyeonwoo Noh, Long
  Ouyang, Cullen O'Keefe, Jakub Pachocki, Alex Paino, Joe Palermo, Ashley
  Pantuliano, Giambattista Parascandolo, Joel Parish, Emy Parparita, Alex
  Passos, Mikhail Pavlov, Andrew Peng, Adam Perelman, Filipe de~Avila
  Belbute~Peres, Michael Petrov, Henrique~Ponde de~Oliveira~Pinto, Michael,
  Pokorny, Michelle Pokrass, Vitchyr~H. Pong, Tolly Powell, Alethea Power,
  Boris Power, Elizabeth Proehl, Raul Puri, Alec Radford, Jack Rae, Aditya
  Ramesh, Cameron Raymond, Francis Real, Kendra Rimbach, Carl Ross, Bob
  Rotsted, Henri Roussez, Nick Ryder, Mario Saltarelli, Ted Sanders, Shibani
  Santurkar, Girish Sastry, Heather Schmidt, David Schnurr, John Schulman,
  Daniel Selsam, Kyla Sheppard, Toki Sherbakov, Jessica Shieh, Sarah Shoker,
  Pranav Shyam, Szymon Sidor, Eric Sigler, Maddie Simens, Jordan Sitkin,
  Katarina Slama, Ian Sohl, Benjamin Sokolowsky, Yang Song, Natalie Staudacher,
  Felipe~Petroski Such, Natalie Summers, Ilya Sutskever, Jie Tang, Nikolas
  Tezak, Madeleine~B. Thompson, Phil Tillet, Amin Tootoonchian, Elizabeth
  Tseng, Preston Tuggle, Nick Turley, Jerry Tworek, Juan Felipe~Cerón Uribe,
  Andrea Vallone, Arun Vijayvergiya, Chelsea Voss, Carroll Wainwright,
  Justin~Jay Wang, Alvin Wang, Ben Wang, Jonathan Ward, Jason Wei, CJ~Weinmann,
  Akila Welihinda, Peter Welinder, Jiayi Weng, Lilian Weng, Matt Wiethoff, Dave
  Willner, Clemens Winter, Samuel Wolrich, Hannah Wong, Lauren Workman, Sherwin
  Wu, Jeff Wu, Michael Wu, Kai Xiao, Tao Xu, Sarah Yoo, Kevin Yu, Qiming Yuan,
  Wojciech Zaremba, Rowan Zellers, Chong Zhang, Marvin Zhang, Shengjia Zhao,
  Tianhao Zheng, Juntang Zhuang, William Zhuk, and Barret Zoph.
\newblock Gpt-4 technical report, 2024.

\bibitem{xie2024travelplanner}
Jian Xie, Kai Zhang, Jiangjie Chen, Tinghui Zhu, Renze Lou, Yuandong Tian,
  Yanghua Xiao, and Yu~Su.
\newblock Travelplanner: A benchmark for real-world planning with language
  agents, 2024.

\bibitem{zheng2024natural}
Huaixiu~Steven Zheng, Swaroop Mishra, Hugh Zhang, Xinyun Chen, Minmin Chen,
  Azade Nova, Le~Hou, Heng-Tze Cheng, Quoc~V. Le, Ed~H. Chi, and Denny Zhou.
\newblock Natural plan: Benchmarking llms on natural language planning, 2024.

\bibitem{multion2024}
{MultiOn AI}.
\newblock Multion ai, 2024.

\bibitem{simular2024}
{Simular AI}.
\newblock Simular ai, 2024.

\bibitem{adept2024}
{Adept AI}.
\newblock Adept ai: Useful general intelligence, 2024.

\bibitem{rabbit2024}
{Rabbit Inc.}
\newblock Rabbit: Your pocket companion, 2024.

\bibitem{humane2024}
{Humane Inc.}
\newblock Humane: Wearable ai, 2024.

\bibitem{limitless2024}
{Limitless AI}.
\newblock Limitless ai, 2024.

\bibitem{xie2024osworld}
Tianbao Xie, Danyang Zhang, Jixuan Chen, Xiaochuan Li, Siheng Zhao, Ruisheng
  Cao, Toh~Jing Hua, Zhoujun Cheng, Dongchan Shin, Fangyu Lei, Yitao Liu,
  Yiheng Xu, Shuyan Zhou, Silvio Savarese, Caiming Xiong, Victor Zhong, and Tao
  Yu.
\newblock Osworld: Benchmarking multimodal agents for open-ended tasks in real
  computer environments, 2024.

\bibitem{bishop2024latent}
William~E Bishop, Alice Li, Christopher Rawles, and Oriana Riva.
\newblock Latent state estimation helps ui agents to reason, 2024.

\bibitem{nakano2022webgpt}
Reiichiro Nakano, Jacob Hilton, Suchir Balaji, Jeff Wu, Long Ouyang, Christina
  Kim, Christopher Hesse, Shantanu Jain, Vineet Kosaraju, William Saunders,
  Xu~Jiang, Karl Cobbe, Tyna Eloundou, Gretchen Krueger, Kevin Button, Matthew
  Knight, Benjamin Chess, and John Schulman.
\newblock Webgpt: Browser-assisted question-answering with human feedback,
  2022.

\bibitem{gur2024realworld}
Izzeddin Gur, Hiroki Furuta, Austin Huang, Mustafa Safdari, Yutaka Matsuo,
  Douglas Eck, and Aleksandra Faust.
\newblock A real-world webagent with planning, long context understanding, and
  program synthesis, 2024.

\bibitem{yu2017survey}
Wei Yu, Fan Liang, Xiaofei He, William~Grant Hatcher, Chao Lu, Jie Lin, and
  Xinyu Yang.
\newblock A survey on the edge computing for the internet of things.
\newblock {\em IEEE access}, 6:6900--6919, 2017.

\bibitem{lin2024awq}
Ji~Lin, Jiaming Tang, Haotian Tang, Shang Yang, Wei-Ming Chen, Wei-Chen Wang,
  Guangxuan Xiao, Xingyu Dang, Chuang Gan, and Song Han.
\newblock Awq: Activation-aware weight quantization for llm compression and
  acceleration, 2024.

\bibitem{alwarafy2020survey}
Abdulmalik Alwarafy, Khaled~A Al-Thelaya, Mohamed Abdallah, Jens Schneider, and
  Mounir Hamdi.
\newblock A survey on security and privacy issues in edge-computing-assisted
  internet of things.
\newblock {\em IEEE Internet of Things Journal}, 8(6):4004--4022, 2020.

\bibitem{ranaweera2021survey}
Pasika Ranaweera, Anca~Delia Jurcut, and Madhusanka Liyanage.
\newblock Survey on multi-access edge computing security and privacy.
\newblock {\em IEEE Communications Surveys \& Tutorials}, 23(2):1078--1124,
  2021.

\bibitem{chen2024octopus2}
Wei Chen and Zhiyuan Li.
\newblock Octopus v2: On-device language model for super agent, 2024.

\bibitem{openai_assistant_overview}
OpenAI.
\newblock Overview of assistants.
\newblock \url{https://platform.openai.com/docs/assistants/overview}, 2024.
\newblock Accessed: 2024-06-23.

\bibitem{yao2023react}
Shunyu Yao, Jeffrey Zhao, Dian Yu, Nan Du, Izhak Shafran, Karthik Narasimhan,
  and Yuan Cao.
\newblock React: Synergizing reasoning and acting in language models, 2023.

\bibitem{shen2024small}
Weizhou Shen, Chenliang Li, Hongzhan Chen, Ming Yan, Xiaojun Quan, Hehong Chen,
  Ji~Zhang, and Fei Huang.
\newblock Small llms are weak tool learners: A multi-llm agent, 2024.

\bibitem{hu2023generalpurpose}
Yafei Hu, Quanting Xie, Vidhi Jain, Jonathan Francis, Jay Patrikar, Nikhil
  Keetha, Seungchan Kim, Yaqi Xie, Tianyi Zhang, Shibo Zhao, Yu~Quan Chong,
  Chen Wang, Katia Sycara, Matthew Johnson-Roberson, Dhruv Batra, Xiaolong
  Wang, Sebastian Scherer, Zsolt Kira, Fei Xia, and Yonatan Bisk.
\newblock Toward general-purpose robots via foundation models: A survey and
  meta-analysis, 2023.

\bibitem{firoozi2023foundation}
Roya Firoozi, Johnathan Tucker, Stephen Tian, Anirudha Majumdar, Jiankai Sun,
  Weiyu Liu, Yuke Zhu, Shuran Song, Ashish Kapoor, Karol Hausman, Brian Ichter,
  Danny Driess, Jiajun Wu, Cewu Lu, and Mac Schwager.
\newblock Foundation models in robotics: Applications, challenges, and the
  future, 2023.

\bibitem{ahn2022i}
Michael Ahn, Anthony Brohan, Noah Brown, Yevgen Chebotar, Omar Cortes, Byron
  David, Chelsea Finn, Chuyuan Fu, Keerthana Gopalakrishnan, Karol Hausman,
  Alex Herzog, Daniel Ho, Jasmine Hsu, Julian Ibarz, Brian Ichter, Alex Irpan,
  Eric Jang, Rosario~Jauregui Ruano, Kyle Jeffrey, Sally Jesmonth, Nikhil~J
  Joshi, Ryan Julian, Dmitry Kalashnikov, Yuheng Kuang, Kuang-Huei Lee, Sergey
  Levine, Yao Lu, Linda Luu, Carolina Parada, Peter Pastor, Jornell Quiambao,
  Kanishka Rao, Jarek Rettinghouse, Diego Reyes, Pierre Sermanet, Nicolas
  Sievers, Clayton Tan, Alexander Toshev, Vincent Vanhoucke, Fei Xia, Ted Xiao,
  Peng Xu, Sichun Xu, Mengyuan Yan, and Andy Zeng.
\newblock Do as i can, not as i say: Grounding language in robotic affordances,
  2022.

\bibitem{du2023video}
Yilun Du, Mengjiao Yang, Pete Florence, Fei Xia, Ayzaan Wahid, Brian Ichter,
  Pierre Sermanet, Tianhe Yu, Pieter Abbeel, Joshua~B. Tenenbaum, Leslie
  Kaelbling, Andy Zeng, and Jonathan Tompson.
\newblock Video language planning, 2023.

\bibitem{lester2021power}
Brian Lester, Rami Al-Rfou, and Noah Constant.
\newblock The power of scale for parameter-efficient prompt tuning.
\newblock {\em arXiv preprint arXiv:2104.08691}, 2021.

\bibitem{li2021prefix}
Xiang~Lisa Li and Percy Liang.
\newblock Prefix-tuning: Optimizing continuous prompts for generation.
\newblock {\em arXiv preprint arXiv:2101.00190}, 2021.

\bibitem{paul2023deep}
Mansheej Paul, Surya Ganguli, and Gintare~Karolina Dziugaite.
\newblock Deep learning on a data diet: Finding important examples early in
  training, 2023.

\bibitem{cao2023instruction}
Yihan Cao, Yanbin Kang, Chi Wang, and Lichao Sun.
\newblock Instruction mining: When data mining meets large language model
  finetuning, 2023.

\bibitem{xia2024less}
Mengzhou Xia, Sadhika Malladi, Suchin Gururangan, Sanjeev Arora, and Danqi
  Chen.
\newblock Less: Selecting influential data for targeted instruction tuning,
  2024.

\bibitem{wang2024diversity}
Peiqi Wang, Yikang Shen, Zhen Guo, Matthew Stallone, Yoon Kim, Polina Golland,
  and Rameswar Panda.
\newblock Diversity measurement and subset selection for instruction tuning
  datasets, 2024.

\bibitem{hu2021lora}
Edward~J. Hu, Yelong Shen, Phillip Wallis, Zeyuan Allen-Zhu, Yuanzhi Li, Shean
  Wang, Lu~Wang, and Weizhu Chen.
\newblock Lora: Low-rank adaptation of large language models, 2021.

\bibitem{wang2023multilora}
Yiming Wang, Yu~Lin, Xiaodong Zeng, and Guannan Zhang.
\newblock Multilora: Democratizing lora for better multi-task learning, 2023.

\bibitem{hayou2024lora}
Soufiane Hayou, Nikhil Ghosh, and Bin Yu.
\newblock Lora+: Efficient low rank adaptation of large models, 2024.

\bibitem{kopiczko2024vera}
Dawid~J. Kopiczko, Tijmen Blankevoort, and Yuki~M. Asano.
\newblock Vera: Vector-based random matrix adaptation, 2024.

\bibitem{zhang2023adalora}
Qingru Zhang, Minshuo Chen, Alexander Bukharin, Nikos Karampatziakis, Pengcheng
  He, Yu~Cheng, Weizhu Chen, and Tuo Zhao.
\newblock Adalora: Adaptive budget allocation for parameter-efficient
  fine-tuning, 2023.

\bibitem{liu2024dora}
Shih-Yang Liu, Chien-Yi Wang, Hongxu Yin, Pavlo Molchanov, Yu-Chiang~Frank
  Wang, Kwang-Ting Cheng, and Min-Hung Chen.
\newblock Dora: Weight-decomposed low-rank adaptation, 2024.

\bibitem{zi2023deltalora}
Bojia Zi, Xianbiao Qi, Lingzhi Wang, Jianan Wang, Kam-Fai Wong, and Lei Zhang.
\newblock Delta-lora: Fine-tuning high-rank parameters with the delta of
  low-rank matrices, 2023.

\end{thebibliography}
\bibliographystyle{unsrt}
% }
\section{Appendix}
\subsection{Function/API description examples}\label{sec:a1}
\begin{lstlisting}[style=mystyle]
def get_trending_news(query, language):
    """
    Retrieves a collection of trending news articles relevant to a specified query and language.

    Parameters:
    - query (str): Topic for news articles.
    - language (str): ISO 639-1 language code. The default language is English ('en'), but it can be set to any valid ISO 639-1 code to accommodate different language preferences (e.g., 'es' for Spanish, 'fr' for French).

    Returns:
    - list[str]: A list of strings, where each string represents a single news article. Each article representation includes the article's title and its URL, allowing users to easily access the full article for detailed information.
    """


def get_weather_forecast(location):
    """
    Provides a weather forecast for a specified location over a given number of days. Each day's forecast includes a brief description of the expected weather conditions.

    Parameters:
    - location (str): The location for which the weather forecast is desired. Can be a city name, ZIP code, or other location identifiers.

    Returns:
    - list[str]: A list of strings, each representing the weather forecast for one day. Each string includes the date and a brief description of the weather conditions. Formatted in 'YYYY-MM-DD: Description' format.
    """


def send_email(recipient, title, content):
    """
    Sends an email to a specified recipient with a given title and content.

    Parameters:
    - recipient (str): The email address of the recipient.
    - title (str): The subject line of the email. This is a brief summary or title of the email's purpose or content.
    - content (str): The main body text of the email. It contains the primary message, information, or content that is intended to be communicated to the recipient.

    Returns:
    """


def search_youtube_videos(query):
    """
    Searches YouTube for videos matching a query.

    Parameters:
    - query (str): Search query.

    Returns:
    - list[str]: A list of strings, each string includes video names and URLs.
    """


def find_route_google_maps(origin, destination, mode):
    """
    Computes a route using Google Maps from an origin to a destination.

    Parameters:
    - origin (str): Starting location.
    - destination (str): Target location.
    - mode (enum): Mode of transportation, options include 'driving', 'walking', 'bicycling', and 'transit'. The default mode is 'driving'.

    Returns:
    - List[str]:  The string provides the route details.
    """


def send_text_message(contact_name, message):
    """
    Sends a text message to the specified contact.

    Parameters:
    - contact_name (str): The name of the recipient contact.
    - message (str): The content of the message to be sent. This is what the recipient will receive.

    Returns:
    """


def create_contact(name, phone_number):
    """
    Creates a new contact entry in the device's address book.

    Parameters:
    - name (str): Full name of the contact. This should include first and last name.
    - phone_number (str): phone number of the contact. The phone number should be provided in a standard format, preferably in E.164 format (e.g., +12345678900 for an international format).

    Returns:
    """


def set_timer_alarm(time, label):
    """
    Sets a timer or alarm for a specified time.

    Parameters:
    - time (str): Alarm time in "HH:MM" 24-hour format. For example, "07:12" for 7:12 AM.
    - label (str): Custom label for the alarm, default is "alarm".

    Returns:
    """


def create_calendar_event(title, start_time, end_time):
    """
    Schedules a new event in the calendar.

    Parameters:
    - title (str): Event title.
    - start_time (str): Event start time as a string in ISO 8601 format "YYYY-MM-DD-HH-MM". For example, "2022-12-31-23-59" for 11:59 PM on December 31, 2022.
    - end_time (str): Event end time as a string in ISO 8601 format "YYYY-MM-DD-HH-MM". Must be after start_time. For example, "2023-01-01-00-00" for 12:00 AM on January 1, 2023.

    Returns:
    """


def set_volume(level, volume_type):
    """
    Sets the volume level for a specified type : "ring" , "media" , "alarm".

    Parameters:
    - level (int): Target volume level, from 0 (mute) to 10 (maximum).
    - volume_type (enum): The category of volume to adjust, select from "ring" , "media" , "alarm".

    Returns:
    """
\end{lstlisting}

%%%%%%%%%%%%%%%%%%%%%%%%%%%%%%%%%%%%%%%%%%%%%%%%%%%%%%%%%%%%%

% \end{CJK*}
\end{document}